\RequirePackage{fix-cm}
\documentclass[smallcondensed]{svjour3}     
\smartqed
\usepackage{graphicx}

\usepackage{latexsym}
\usepackage{graphicx}
\RequirePackage{natbib}
\RequirePackage{url}

\usepackage{amsmath}
\usepackage{amsfonts}
\usepackage{bm}
\usepackage{url}
\usepackage{indentfirst}

\def\argmax{{\rm argmax}}
\def\argmin[{{\rm argmin}}

\begin{document}

\title{Online Aggregation of Unbounded Losses Using Shifting Experts with
Confidence\thanks{This paper is an extended version of the conference paper~\citet{Vyu2017}. This work was
supported by Russian Science Foundation, project 14-50-00150.}
}


\author{Vladimir V'yugin \and Vladimir Trunov}


\institute{Vladimir V'yugin \at
              Institute for Information Transmission Problems, Moscow \\
              \email{vyugin@iitp.ru}
           \and
           Vladimir Trunov \at
              Institute for Information Transmission Problems, Moscow \\
              \email{trunov@iitp.ru}
              }

\date{Received: date / Accepted: date}

\maketitle

\def\K{{\rm K}}
\def\P{{\rm P}}
\def\Rand{{\rm Rand}}
\def\Max{{\rm Max}}
\def\argmax{{\rm argmax}}
\def\Dev{{\rm Dev}}
\def\argmin{{\rm argmin}}
\def\Fluc{{\rm fluc}}

\def\argmax{{\rm argmax}}
\def\Dev{{\rm Dev}}
\def\argmin[{{\rm argmin}}
\def\Var{{\rm Var}}
\def\x{{\bf x}}
\def\y{{\bf y}}
\def\g{{\bf g}}
\def\s{{\bf s}}
\def\p{{\bf p}}
\def\q{{\bf q}}
\def\z{{\bf z}}
\def\w{{\bf w}}
\def\l{{\bf l}}
\def\0{{\bf 0}}
\def\c{{\bf c}}
\def\Z{{\bf Z}}
\def\b{{\bf b}}
\def\a{{\bf a}}
\def\X{{\bf X}}
\def\e{{\bf e}}
\def\Pr{{\rm Pr}}
\def\C{{\bf C}}
\def\0{{\bf 0}}
\def\f{{\bf f}}
\def\I{{\bf I}}
\def\P{{\bf P}}
\def\Q{{\bf Q}}




\begin{abstract}
We develop the setting of sequential prediction based on
shifting experts and on a ``smooth'' version of the method of specialized experts.
To aggregate experts predictions, we use the AdaHedge algorithm, which is a version of
the Hedge algorithm with adaptive learning rate, and extend it by the meta-algorithm
Fixed Share. Due to this, we combine the advantages of both algorithms:
(1) we use the shifting regret which is a more optimal characteristic of the algorithm;
(2) regret bounds are valid in the case of signed unbounded losses of the experts.
Also, (3) we incorporate in this scheme a ``smooth'' version of the method of
specialized experts which allows us to make more flexible and accurate predictions.
All results are obtained in the adversarial setting -- no assumptions are made
about the nature of data source.
We present results of numerical experiments for short-term forecasting of electricity
consumption based on a real data.
\end{abstract}

\section{Introduction}

We consider sequential prediction in the general framework of Decision Theoretic Online
Learning or the Hedge setting by~\citet{FrS97}, which is a variant of
prediction with expert advice, see e.g. (\citealt{LiW94},~\citealt{FrS97},
~\citealt{Vov90},~\citealt{VoV98},~\citealt{cesa-bianchi}).

The aggregating algorithm updates the experts weights at the end of each trial
using losses suffered by the experts in the past.
In classical setting (\citealt{FrS97},~\citealt{Vov90}),
the process of an expert $i$ weights updating is based on exponential
weighting with a constant or variable learning rate $\eta$:
\begin{eqnarray}
w_{i,t+1}=\frac{w_{i,t}e^{-\eta l_{i,t}}}{\sum\limits_{j=1}^N w_{j,t}e^{-\eta l_{j,t}}},
\label{weights-up-1}
\end{eqnarray}
where $l_{i,t}$ is the loss suffered by the expert $i$ at step $t$.

The goal of the algorithm is to design weight updates that guarantee that the loss
of the aggregating algorithm is never much larger than the loss of the best expert
or the best convex combination of the losses of the experts.

So, here the best expert or a convex combination of experts serves as a comparator.
By a comparison vector we mean a vector $\q=(q_1,\dots ,q_N)$ such that
$q_1+\dots +q_N=1$ and all its components are nonnegative. We compare
the cumulative loss of the aggregating algorithm and a convex combination of
the losses $\sum\limits_{t=1}^T (\q\cdot \l_t)$,
where $\l_t=(l_{1,t},\dots ,l_{N,t})$ is a vector containing the losses of
the experts at time $t$.

A more challenging goal is to learn well when the comparator $\q$ changes over time,
i.e. the algorithm competes with the cumulative sum $\sum\limits_{t=1}^T (\q_t\cdot \l_t)$,
where comparison vector $\q_t$ changes over time. An important special case
is when $\q_t$ are unit vectors, then the sequence of trials is partitioned
into segments. In each segment the loss of the algorithm
is compared to the loss of a particular expert and this expert changes at the beginning of
a new segment. The goal of the aggregation algorithm is to do almost as well as the
sum of losses of experts forming the best partition.
Algorithms and bounds for shifting comparators were presented by~\citet{HeW98}.
This method called Fixed Share was generalized by~\citet{BoW2002} to the method of
Mixing Past Posteriors (MPP) in which arbitrary mixing schemes are considered.
In what follows, MPP mixing schemes will be used in our algorithms.

Most papers in the prediction with expert advice setting either consider 
uniformly bounded losses or assume the existence of a specific loss function
(see~\citealt{Vov90},~\citealt{cesa-bianchi}).
But in some practical applications, this assumption is too restrictive.
We allow losses at any step to be unbounded and signed.
The notion of a specific loss function is not used.

AdaHedge presented by~\citet{follow} is among a few algorithms
that do not have similar restrictions. This algorithm is a version of the classical
Hedge algorithm of~\citet{FrS97} and is a refinement of the~\citet{cesa-bianchi} algorithm.
AdaHedge is completely parameterless and tunes the learning rate $\eta$
in terms of a direct measure of past performance.

In~\citet{follow}, an upper bound for regret of this algorithm is presented
which is free from boundness assumptions for losses of the experts:
\begin{equation}
R_T\le 2\sqrt{S_T\frac{(L_T^*-L_T^-)(L_T^+-L_T^*)}{L_T^+-L_T^-} \ln N} +
\left(\frac{16}{3} \ln N + 2\right)S_T,
\label{Regret_AH}
\end{equation}
where $L^*_T$ is the loss of the best expert,
for other notations see Table~\ref{tab-1} below.

In the case where losses of the experts are uniformly 
bounded the upper bound (\ref{Regret_AH}) takes the form $O(\sqrt{T\ln N})$.

We emphasize that the versions of Fixed Share and MPP algorithms presented
by~\citet{HeW98} and~\citet{BoW2002} use a constant learning rate,
while the AdaHedge uses adaptive learning rate which is tuned on-line.

The first contribution of this paper is that we present the ConfHedge-1 algorithm
which combines advantages of both these algorithms:
(1) we use the shifting regret which is a more optimal characteristic
of the algorithm; (2) regret bounds are valid in the case of signed unbounded losses
of the experts.

The application we will consider below is the sequential short-term (one-hour-ahead)
forecasting of electricity consumption will take place in a variant of the basic
problem of prediction with expert advice called prediction with specialized (or
sleeping) experts. At each round only some of the experts output a prediction
while the other ones are inactive. Each expert is
expected to provide accurate forecasts mostly in given external conditions, that
can be known beforehand. For instance, in the case of the prediction of electricity
consumption, experts can be specialized to a season, temperature, to working days
or to public holidays, etc.

The method of specialized experts was first proposed by~\citet{Freu-2} and
further developed by~\citet{adams},~\citet{ChV2009},~\citet{electricity},~\citet{KAS2015}.
With this approach, at each step $t$, a set of specialized experts
$E_t\subseteq\{1,\dots, N\}$ is given. A specialized expert $i$ issues its forecasts
not at all steps $t=1,2,\dots$, but only when $i\in E_t$. At any step, the aggregating
algorithm uses forecasts of only ``active (non-sleeping)'' experts.

The second contribution of this paper is that we have incorporated into
ConfHedge-1 a smooth generalization of the method of specialized experts.
At each time moment $t$, we complement the expert $i$ forecast by a confidence level
which is a real number $p_{i,t}\in [0,1]$.

The setting of prediction with experts that report their confidences as a number
in the interval $[0,1]$ was first studied by~\cite{BlM2007} and further developed
by~\cite{CBMS2007}, ~\citet{GGS2011},~\citet{GSE2014}.

In particular, $p_{i,t}=1$ means that the expert forecast is used in full, whereas
in the case of $p_{i,t}=0$ it is not taken into account at all (the expert sleeps).
In cases where $0<p_{i,t}<1$ the expert's forecast is partially taken into account.
For example, with a gradual drop in temperature a corresponded specialized expert
gradually loses its ability for accurate predictions of electricity consumption.
The dependence of $p_{i,t} $ on values of exogenous parameters can be predetermined
by a specialist in the domain or can be constructed using regression analysis on
historical data.

In Section~\ref{general-loss-1}, we present the ConfHedge-1 algorithm, which
is a loss allocation algorithm adapted for the case, where the losses of the
experts can be signed and unbounded. Also, this algorithm takes into account
the confidence levels of the experts predictions.
In Section~\ref{general-loss-2}, ConfHedge-2 variant
of this algorithm is presented for the case when experts make forecasts
and calculate their losses using a convex loss function.

In Theorem~\ref{main-result} we present the upper
bounds for the shifting regret of these algorithms.
The proof of this theorem is given in Section~\ref{tech-det-1}.
Some details of the proof from~\citet{follow} are presented as a
supplementary material in Section~\ref{sec-3}.
All results are obtained in the adversarial setting and
no assumptions are made about the nature of data source.

In Section~\ref{subsec-2}, the techniques of confidence level selection and experts
training are presented. We also present the results of numerical experiments
of the short-term prediction of electricity consumption with the use of the
proposed algorithms.

The approach that sets the confidence levels for expert predictions
of electricity consumption is more general than the approach used in
the paper~\citet{electricity}, which uses ``sleeping'' experts. In our numerical
experiments the aggregating algorithm with soft confidence levels outperforms
other versions of aggregating algorithms including ones which use sleeping experts.

\begin{table}\label{tab-1}
$N$ -- number of experts

$\l_t=(l_{1,t},\dots , l_{N,t})$ -- loss vector at step $t$

$\p_t=(p_{1,t},\dots ,p_{N,t})$ -- vector of confidences at step $t$

$\hat\l_t=(\hat l_{1,t},\dots , \hat l_{N,t})$ -- vector of transformed losses

$l_t^-=\min_{1\le i\le N}l_{i,t}$, $l_t^+=\max_{1\le i\le N}l_{i,t}$ -- min and max loss

$s_t=l_t^+-l_t^-$ -- loss range

$\q_t=(q_{1,t},\dots ,q_{N,t})$ -- comparison vector at step $t$

$\w^{\mu}_t=(w^\mu_{1,t},\dots ,w^\mu_{N,t})$ -- experts weights

$\w_t=(w_{1,t},\dots ,w_{N,t})$ -- experts posterior weights at step $t$

$\w^*_t=(w^*_{1,t},\dots ,w^*_{N,t})$ -- the learner prediction, where
$w^*_{i,t}=\frac{w_{i,t}p_{i,t}}{\sum_{i=1}^N w_{i,t}p_{i,t}}$ for $1\le i\le N$.

$h_t=(\w^*_i\cdot l_i)$ -- Hedge loss (dot product of two vectors).

$m_t=-\frac{1}{\eta_t}\sum\limits_{i=1}^N w_{i,t}e^{-\eta_t \hat l_{i,t}}$ -- mixloss

$\delta_t=h_t-m_t$ -- mixability gap

$\alpha_t$ -- Fixed Share parameter (we put $\alpha_t=\frac{1}{t}$)

$L_T^-=\sum\limits_{t=1}^T l_t^-$, $L_T^+=\sum\limits_{t=1}^T l_t^+$ -- cumulative
minimal and maximal losses

$S_T=\max_{1\le t\le T} s_t$ -- maximum loss range

$H_T=\sum\limits_{t=1}^T h_t$ -- algorithm cumulative loss

$M_T=\sum\limits_{t=1}^T m_t$ -- cumulative mixloss

$\Delta_T=\sum_{t=1}^T\delta_t$ -- cumulative gap

$R^{(\q)}_T=\sum\limits_{t=1}^T\sum\limits_{i=1}^N q_{i,t} p_{i,t} (h_t-l_{i,t})$ --
confidence shifting regret

$\eta_t=\frac{\ln^*N}{\Delta_{t-1}}$ -- variable learning rate, 
where $\ln^*N=\max\{1,\ln N\}$ 

\caption{Basic notations and definitions}

\end{table}

\section{Online loss allocation algorithm}\label{general-loss-1}

In this section we present an algorithm for the optimal online allocation
of unbounded signed losses of the experts. In Section~\ref{general-loss-2}, a variant
of this algorithm will be presented for the case when experts make forecasts
and calculate their losses using a convex loss function.

We assume that at each step $t$, along with the losses $l_{i,t}$ of experts,
theirs confidence levels are given -- a vector
$\p_t=(p_{1,t},\dots ,p_{N,t})$, where $p_{i,t}\in [0,1]$ for $1\le i\le N$.
We assume that $\|\p_t\|_1>0$ for all $t$.

We can interpret the number $ p_{i,t} $ as the algorithm's internal probability
of following the expert $i$ prediction. In this case, we define the auxiliary
virtual losses of the expert as a random variable
\[
\tilde l_{i,t}=
\left\{
    \begin{array}{l}
      l_{i,t} \mbox{ with probability } p_{i,t},
    \\
      h_t \mbox{ with probability } 1-p_{i,t},
    \end{array}
  \right.
\]
where $h_t$ is the aggregating algorithm loss. Denote
$\hat l_{i,t}=E_{\p_t}[\tilde l_{i,t}]=p_{i,t} l_{i,t}+(1-p_{i,t})h_t$
the mathematical expectation of a virtual loss of an expert $i$ with respect to
the probability distribution $\p_{i,t}=(p_{i,t},1-p_{i,t})$.

At any step $t$ we use cumulative weights $w_{i,t}$ of the experts $1\le i\le N$
which were computed at the previous step.
The algorithm loss is defined as $h_t=\sum\limits_{i=1}^N w_{i,t}\hat l_{i,t}$.

These definitions contain a logical circle -- virtual losses
are determined through loss of the algorithm, and the latter is determined through
virtual losses. Nevertheless, all these quantities can be effectively calculated
using the fixed-point method proposed by~\citet{ChV2009}. We have
\begin{eqnarray*}
h_t=\sum_{i=1}^N w_{i,t}\hat l_{i,t}=
\sum_{i=1}^N w_{i,t}(p_{i,t}l_{i,t}+(1-p_{i,t})h_t)=
\sum_{i=1}^N w_{i,t} p_{i,t}(l_{i,t}-h_t)+h_t.
\end{eqnarray*}
Canceling out the identical terms on the left and on the right sides, we obtain
expression for calculating $h_t$:
\begin{eqnarray}
h_t=\frac{
\sum_{i=1}^N w_{i,t}p_{i,t} l_{i,t}
}
{\sum_{i=1}^N w_{i,t}p_{i,t}
}
\label{mixloss-2}
\end{eqnarray}
After the value of $h_t$ was calculated by the formula (\ref{mixloss-2}),
the weights can be calculated as
\begin{eqnarray}
w^\mu_{i,t}=\frac{
w_{i,t} e^{-\eta\hat l_{i,t}}
}
{
\sum_{s=1}^N w_{s,t} e^{-\eta\hat l_{s,t}}
}=
\frac{
w_{i,t} e^{-\eta p_{i,t}(l_{i,t}-h_t)}
}
{
\sum_{s=1}^N w_{s,t} e^{-\eta p_{s,t}(l_{s,t}-h_t)}
}
\label{weight-update-rule}
\end{eqnarray}
using the known value $h_t$.\footnote{In the simple Hedge we put $w_{i,t+1}=w^\mu_{i,t}$.
Some other mixing schemes will be given below.}
Also, we compute the mixloss
$
m_t=-\frac{1}{\eta_t}\sum\limits_{i=1}^N w_{i,t}e^{-\eta_t\hat l_{i,t}}
$
and the mixability gap $\delta_t=h_t-m_t$, which are used in the construction
of the algorithm.

By the method MPP of~\citet{BoW2002}, a mixing scheme is defined
by a vector $\beta^{t+1}=(\beta^{t+1}_0,\dots, \beta^{t+1}_t)$,
where $\sum\limits_{s=0}^t\beta^{t+1}_s=1$ and $\beta^{t+1}_s\ge 0$ for $0\le s\le t$.

In what follows the vector $\w^\mu_t=(w^\mu_{1,t},\dots ,w^\mu_{N,t})$
presents the normalized experts weights at step $t$. The corresponding posterior
probability distribution $\w_{t+1}=(w_{1,t+1},\dots ,w_{N,t+1})$ for step $t+1$
is defined as a convex combination
$\w_{t+1}=\sum\limits_{s=0}^t\beta^{t+1}_s w^\mu_s$ with weights $\beta^{t+1}_s$,
$0\le s\le t$, where $\w^\mu_s=(w^\mu_{1,s},\dots ,w^\mu_{N,s})$.

The vector $\beta^{t + 1}$ defines the weights by which the past distributions
of experts are mixed. It can be re-set at each step $t$.

The ConfHedge-1 algorithm for mixing the posteriori distributions of experts is
given below. Unlike standard exponential mixing algorithms, this algorithm uses
not only the current accumulated weights of experts, but also mixes these weights
and all the weights accumulated in past steps.

\smallskip

\textbf{\bf ConfHedge-1}

{\small
\medskip\hrule\hrule\medskip

\smallskip
\noindent Put $w_{i,1}=w^\mu_{i,0}=\frac{1}{N}$ for $i=1,\dots , N$, $\Delta_0=0$,
$\eta_1=\infty$.

\noindent FOR $t=1,\dots ,T$
\\
Receive confidence levels $\p_t=(p_{1,t},\dots ,p_{N,t})$ of the experts $1\le i\le N$,
where $\|\p_t\|_1>0$.
\\
Predict with the distribution $\w^*_t=(w^*_{1,t},\dots ,w^*_{N,t})$, where
$w^*_{i,t}=\frac{w_{i,t}p_{i,t}}{\sum_{i=1}^N w_{i,t}p_{i,t}}$ for $1\le i\le N$.
\\
Receive a vector $\l_t=(l_{1,t},\dots ,l_{N,t})$ containing the losses of the experts.
\\
Compute the loss $h_t=(\l_t\cdot\w^*_t)$
of the algorithm.
\\
Update the weights and the learning parameter in three stages:
\\
\textbf{Loss Update}
\\
Define $w^\mu_{i,t}=\frac{w_{i,t}e^{-\eta_t p_{i,t}(l_t^i-h_t)}}
{\sum\limits_{s=1}^N w_{s,t}e^{-\eta_t p_{s,t}(l_{s,t}-h_t)}}$ for $1\le i\le N$.
\\
\textbf{Mixing Update}
\\
Choose a mixing scheme $\beta^{t+1}=(\beta^{t+1}_0,\dots ,\beta^{t+1}_t)$ and define
future weights of the experts
\\
$w_{i,t+1}=\sum\limits_{s=0}^t\beta^{t+1}_s w^\mu_{i,s}$ for $1\le i\le N$.
\\
\textbf{Learning Parameter Update}
\\
Define mixloss
$m_t=-\frac{1}{\eta_t}\ln\sum_{i=1}^N w_{i,t} e^{-\eta_t(p_{i,t}l_{i,t}+(1-p_{i,t})h_t)}$.
Let $\delta_t=h_t-m_t$ and $\Delta_t=\Delta_{t-1}+\delta_t$.
Define the learning rate $\eta_{t+1}=\ln^*N/\Delta_t$ for use at the next
step $t+1$.
\\
\noindent ENDFOR

\smallskip

\medskip\hrule\hrule\medskip
}
\smallskip

We have $m_t\le h_t$ by convexity of the exponent,
then $\delta_t\ge 0$ and $\Delta_t\le\Delta_{t+1}$ for all $t$.


We will use the following mixing schemes by~\citet{BoW2002}:

\textbf{Example 1.} A version of Fixed Share by~\citet{HeW98}
(see also~\citealt{cesa-bianchi},~\citealt{Vov99}) with a variable
learning rate is defined by the following mixing scheme. Let a sequence
$1\ge\alpha_1\ge\alpha_2\ge\dots >0$ of parameters be given.
Define $\beta^{t+1}_t=1-\alpha_{t+1}$ and
$\beta^{t+1}_0=\alpha_{t+1}$ ($\beta^{t+1}_s=0$ for $0<s<t$).
The corresponding prediction for step $t+1$ is defined
$$
w_{i,t+1}=\frac{\alpha_{t+1}}{N}+(1-\alpha_{t+1})w^\mu_{i,t}
$$
for all $1\le i\le N$. In what follows we put $\alpha_t=1/t$ for all $t$.

\textbf{Example 2.} Uniform Past by~\citet{BoW2002}
with a variable learning rate. Put $\beta^{t+1}_t=1-\alpha_{t+1}$ and
$\beta^{t+1}_s=\frac{\alpha_{t+1}}{t}$ for $0\le s<t$.
The corresponding prediction for step $t+1$ is defined
$$
w_{i,t+1}=\alpha_{t+1}\sum\limits_{s=0}^{t-1}\frac{w^\mu_{i,s}}{t}+(1-\alpha_{t+1})w^\mu_{i,t}
$$
for all $i$ and $t$.


\citet{BoW2002} considered the notion of shifting regret with respect
to a sequence $\q_1,\q_2,\dots,\q_T$ of comparison vectors:
$R_T=H_T-\sum\limits_{t=1}^T (\q_t\cdot\l_t)$.\footnote{The notion of regret
with respect to a comparison vector was first defined by~\citet{KiW99}.}

In the presence of confidence values, we consider the corresponding confidence
shifting regret $R^{(\q)}_T=H_T-L^{(\q)}_T$, where
$$
L^{(\q)}_T=\sum\limits_{t=1}^T (\q_t\cdot\hat\l_t)=
\sum\limits_{t=1}^T\sum_{i=1}^N q_{i,t}\hat l_{i,t}=
\sum_{t=1}^T\sum_{i=1}^N q_{i,t}(p_{i,t}l_{i,t}+(1-p_{i,t})h_t)
$$
and $\hat\l_t=(\hat l_{1,t},\dots ,\hat l_{N,t})$ and $\q_t=(q_{1,t},\dots ,q_{N,t})$
is a comparison vector at step $t$.

By definition this regret can be represented as
$$
R^{(\q)}_T=\sum\limits_{t=1}^T\sum\limits_{i=1}^N q_{i,t} p_{i,t} (h_t-l_{i,t}).
$$
If $p_{i,t}=1$ for all $i$ and $t$ then $R^{(\q)}_T=R_T$.

The quantity $L^{(\q)}_T$ depends on $h_t$. To avoid this dependence, we will consider
its lower and upper bounds:
$L^{(\q-)}_T\le L^{(\q)}_T\le L^{(\q+)}_T$, where
\begin{eqnarray}
L^{(\q-)}_T=\sum_{t=1}^T\sum_{i=1}^N q_{i,t}(p_{i,t}l_{i,t}+(1-p_{i,t})l^-_t),
\nonumber
\\
L^{(\q+)}_T=\sum_{t=1}^T\sum_{i=1}^N q_{i,t}(p_{i,t}l_{i,t}+(1-p_{i,t})l^+_t).
\label{l-q-bounds-1}
\end{eqnarray}

Assume that the losses of the experts are bounded, for example, $l_{i,t}\in [0,1]$
for all $i$ and $t$. Using the techniques of Section~\ref{tech-det-1}
for $\eta_t\sim\sqrt{\frac{\ln^*N}{t}}$, we can prove that
\begin{eqnarray}
R^{(\q)}_T=O\left((k+1)\left(\ln T\sqrt{T}+\sqrt{T\ln^*N}\right)\right).
\label{bound-loss-1}
\end{eqnarray}
where $k$ is the number of switches of comparison vectors $\q_t$
on the time interval $1\le t\le T$.\footnote{Does this bound is tight is an
open question. Some lower bounds for mixloss (for the logarithmic loss function
with the learning rate $\eta=1$) were obtained by~\citet{adams}. They
show an information-theoretic lower bound for mixloss that must
hold for any algorithm, and which is tight for Fixed Share.}

Our goal is to obtain a similar bound in the absence of boundness assumptions
for the expert losses.

The following theorem presents the upper bounds for the confidence shifting regret
in the case where no assumptions are made about boundness of the losses of the experts.
\begin{theorem}\label{main-result}
Let the mixing scheme from Example 1 was used. Then for any $T$ and for any 
sequence $\q_1,\dots ,\q_T$ of comparison vectors,
\begin{eqnarray}
R^{(\q)}_T\le
\frac{1}{2}\gamma_{k,T}\sqrt{\sum\limits_{t=1}^T s_t^2\ln^*N}+
\gamma_{k,T}\left(\frac{2}{3}\ln^*N+1\right)S_T,
\label{main-result-1aa-4}
\\
R^{(\q)}_T\le\gamma_{k,T}\sqrt{S_T\frac{(L^+_T-L^{(\q-)}_T)
(L^{(\q+)}_T-L^-_T)}{L^+_T-L^-_T}\ln^* N}+~
\nonumber
\\
\gamma_{k,T}\left(\left(\gamma_{k,T}+\frac{2}{3}\right)\ln^*N+1\right)S_T,
\label{main-result-1a}
\end{eqnarray}
where $\gamma_{k,T}=(k+2)(\ln T+1)$ and $k$ is the number of switches of the comparison
vectors on the time interval $1\le t\le T$.
\end{theorem}
The bound (\ref{main-result-1aa-4}) is an analogue for the shifting experts
of the bound from~\citet{CBMS2007} and the bound (\ref{main-result-1a})
is an analogue of the bound (16) of Theorem 8 from~\citet{follow}.
Proof of Theorem~\ref{main-result} is given in Sections~\ref{tech-det-1} and~\ref{sec-3}.

A disadvantage of the bounds (\ref{main-result-1a}) and (\ref{main-result-1aa-1}) below
is in the presence of a term that depends quadratically on the
number $k$ of switches. Whether such a dependence is necessary is an open question.
However, this term does not depend on the loss of the algorithm,
it has only a slowly growing multiplicative factor $O(\ln^2 T)$.
Corollary~\ref{main-result-1aw} below shows that in some special cases this dependence
can be eliminated.

The bound (\ref{main-result-1a}) of Theorem~\ref{main-result} can be simplified
in the different ways:
\begin{corollary}\label{main-result-1aw}
For any $T$ and for any sequence $\q_1,\dots ,\q_T$ of comparison vectors,
\begin{eqnarray}
R^{(\q)}_T\le\gamma_{k,T}\sqrt{S_T(L^{(\q+)}_T-L^-_T)\ln^*N}+
\gamma_{k,T}\left(\left(\gamma_{k,T}+\frac{2}{3}\right)\ln^*N+1\right)S_T,
\label{main-result-1aa-1}
\\
R^{(\q)}_T\le\gamma_{k,T}\sqrt{S_T(L^+_T-L^{(\q-)}_T)\ln^*N}+
\gamma_{k,T}\left(\frac{2}{3}\ln^*N+1\right)S_T,~~~~~~~~~~~~
\label{main-result-1aa-3}
\\
R^{(\q)}_T\le
\gamma_{k,T}\sqrt{S_T(L^+_T-L^-_T)\ln^*N}+
\gamma_{k,T}\left(\frac{2}{3}\ln^*N+1\right)S_T.~~~~~~~~~~~~~~~~
\label{main-result-1aa-2}
\end{eqnarray}
\end{corollary}

The bound (\ref{main-result-1aa-3}) linearly depends on the number of switches
(for the proof see Section~\ref{sec-3}).
The bound (\ref{main-result-1aa-2}) follows from (\ref{main-result-1aa-3}).
If the losses of the experts are uniformly bounded then the bound (\ref{main-result-1aa-2})
is of the same order as the bound (\ref{bound-loss-1}).

An important special case of Theorem~\ref{main-result} is when the comparison
vectors $\q_t=\e_{i_t}$ are unit vectors
and $p_{i,t}\in\{0,1\}$, i.e., the specialists case is considered for composite
experts $i_1,\dots ,i_T$. Then the confidence shifting regret equals
$R^{(\q)}_T=\sum\limits_{t:p_{i_t,t}=1}^T (h_t-l_{i_t,t})$ and the corresponding
differences in the right-hand side of inequality (\ref{main-result-1a}) are
$L^+_T-L^{(\q-)}_T=\sum\limits_{t:p_{i_t,t}=1}^T (l^+_t-l_{i_t,t})+
\sum\limits_{t:p_{i_t,t}=0}^T s_t$ and
$L^{(\q+)}_T-L^-_T=\sum\limits_{t:p_{i_t,t}=1}^T (l_{i_t,t}-l^-_t)+
\sum\limits_{t:p_{i_t,t}=0}^T s_t$.

The bound (\ref{main-result-1aa-3}) is important if the algorithm is to be used
for a scenario in which we are
provided with a sequence of gain vectors $\g_t$ rather than losses: we can transform these
gains into losses using $\l_t=-\g_t$, and then run the algorithm. Assume that
$p_{i,t}=1$ for all $i$ and $t$. The bound then implies
that we incur small regret with respect to a composite expert if it has very small
cumulative gain relative to the minimum gain (see also~\citealt{follow}).

The similar bounds for the mixing scheme of Example 2 also can be obtained, where
$\gamma_{k,T}=(2k+3)\ln T+(k+2)$ (see Section~\ref{tech-det-1}).

\section{Numerical experiments}\label{sec-4}

Section~\ref{subsec-1} presents the results of applying ConfHedge-1
to synthetic data. In Section~\ref{subsec-2} the results
of the short-term prediction of electricity consumption are presented. We use in
these experiments the ConfHedge-2 algorithm which is a variant the previous
algorithm adapted for the case, where experts present the numerical forecasts.
The scheme of this algorithm is given in Section~\ref{general-loss-2}.

\subsection{Unbounded signed losses}\label{subsec-1}

\begin{figure}[!htb]
\centering\includegraphics[scale=0.45]
        {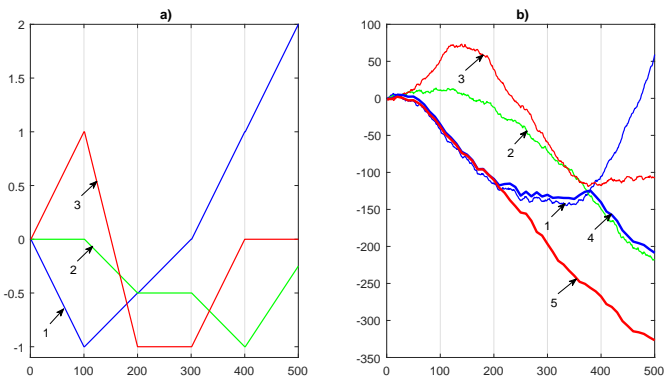}
        \caption{{\small Results of the experiment on the synthetic data.
Left subfigure (a) shows the mean values of one-step experts losses (lines 1, 2, and 3).
Right subfigure (b) shows
cumulative losses of individual experts (thin lines 1, 2 and 3),
AdaHedge and ConfHedge-1 cumulative losses (thick lines 4 and 5)}}\label{fig-1}
                  \end{figure}

The first experiment was performed on synthetic data, where one-step losses of
experts are signed, unbounded and perturbed by $N(0,1)$ additive noise.
Confidence levels of all experts are always equal to one.
Figure~\ref{fig-1}a shows mean values of these one-step expert losses.
Figure~\ref{fig-1}b shows cumulative losses of three individual experts and
cumulative losses of AdaHedge and ConfHedge-1. These experiments show
that ConfHedge-1 is non-inferior to AdaHedge, and, after some time, even outperforms it.

\subsection{Aggregation of expert forecasts}\label{general-loss-2}

In this section we suppose that the losses of the experts are computed using a convex
in $\gamma$ loss function $\lambda(\omega,\gamma)$, where $\omega$ is an outcome
and $\gamma$ is a forecast. Outcomes can belong to an arbitrary set, forecasts form
a linear space.\footnote{
In our experiments, the absolute loss function $\lambda(\omega,\gamma)=|\omega-\gamma|$
was used, where $\omega$ and $\gamma$ are real numbers.
In practical applications, we can also use its biased variant
$\lambda(\omega,\gamma)=\mu_1|\omega-\gamma|_{-}+\mu_2|\omega-\gamma|_{+}$,
where $|r|_{-}=-\min\{0,r\}$ and $|r|_{+}=\max\{0,r\}$.
The positive numbers $\mu_1$ and $\mu_2$ provide a balance of losses between
the deviations of the forecasts $\gamma$ and outcomes $\omega$
in the positive and negative directions.}

Let at any step $t$ the experts forecasts $\c_t=(c_{1_t},\dots ,c_{N,t})$
and their confidence levels $\p_t=(p_{1,t},\dots ,p_{N,t})$ are given. Here
$p_{i,t}\in [0,1]$ for all $1\le i\le N$. Define the auxiliary virtual
experts forecasts
\[
\tilde c_{i,t}=
\left\{
    \begin{array}{l}
      c_{i,t} \mbox{ with probability } p_{i,t},
    \\
      \gamma_t \mbox{ with probability } 1-p_{i,t},
    \end{array}
  \right.
\]
where $\gamma_t$ is a forecast of the aggregating algorithm.
Then the mathematical expectation of any expert $i$ forecast is equal to
$\hat\c_{i,t}=E_{\p_{i,t}}[\tilde c_{i,t}]=p_{i,t}c_{i,t}+(1-p_{i,t})\gamma_t$.

Define the aggregating algorithm forecast
\begin{eqnarray}\label{for-1}
\gamma_t=\sum_{i=1}^N w_{i,t}\hat c_{i,t}.
\end{eqnarray}
In order to get rid of the logical circle in these definitions,
we use the fixed point method by~\citet{ChV2009}. We have
\begin{eqnarray*}
\gamma_t=\sum_{i=1}^N w_{i,t}\hat c_{i,t}=
\sum_{i=1}^N w_{i,t}(p_{i,t} c_{i,t}+(1-p_{i,t})\gamma_t)=
\sum_{i=1}^N w_{i,t} p_{i,t}(c_{i,t}-\gamma_t)+\gamma_t.
\end{eqnarray*}
Cancel the same terms on the left and on the right sides and obtain
\begin{eqnarray}
\gamma_t=\frac{\sum_{i=1}^N p_{i,t}w_{i,t} c_{i,t}}{\sum_{i=1}^N p_{i,t}w_{i,t}}.
\label{pred-2}
\end{eqnarray}
The further calculations are given in the scheme of ConfHedge-2 below.

\smallskip

{\bf ConfHedge-2}

{\small
\medskip\hrule\hrule\medskip
\smallskip

\noindent Define $w_{i,1}=w^\mu_{i,0}=\frac{1}{N}$ for $i=1,\dots , N$, $\Delta_0=0$,
$\eta_1=\infty$.

\noindent FOR $t=1,\dots ,T$
\\
Receive the expert forecasts $\c_t=(c_{1,t},\dots ,c_{N,t})$ and and their
confidence levels $\p_t=(p_{1,t},\dots ,p_{N,t})$.
\\
Compute the aggregating algorithm forecast
$\gamma_t=\frac{\sum_{i=1}^N p_{i,t}w_{i,t} c_{i,t}}{\sum_{i=1}^N p_{i,t}w_{i,t}}$.
\\
Receive an outcome $\omega_t$ and compute the experts losses
$\l_t=(l_{1,t},\dots ,l_{N,t})$, where $l_{i,t}=\lambda(\omega_t,c_{i,t})$, $1\le i\le N$,
and the algorithm loss $a_t=\lambda(\omega_t,\gamma_t)$.
\\
Update experts weights and learning parameter in three stages:
\\
{\bf Loss Update}
\\
Define
\\
$w^\mu_{i,t}=\frac{w_{i,t}e^{-\eta_t p_{i,t}(l_{i,t}-a_t)}}
{\sum\limits_{s=1}^N w_{s,t}e^{-\eta_t p_{s,t}(l_{s,t}-a_t)}}$ for $1\le i\le N$.
\\
{\bf Mixing Update}
\\
Define future experts weights
$w_{i,t+1}=\frac{\alpha_{t+1}}{N}+(1-\alpha_{t+1})w^\mu_{i,t}$
for $1\le i\le N$, where $\alpha_t=\frac{1}{t}$.
\\
{\bf Learning Parameter Update}
\\
Compute the mixloss
\\
$m_t=-\frac{1}{\eta_t}\ln\sum_{i=1}^N w_{i,t} e^{-\eta_t(p_{i,t}l_{i,t})+(1-p_{i,t})a_t)}$.
\\
Define $\delta_t=h_t-m_t$, where $h_t=\sum_{i=1}^N w_{i,t}(p_{i,t}l_{i,t}+(1-p_{i,t})a_t)$,
define also $\Delta_t=\Delta_{t-1}+\delta_t$.
After that set $\eta_{t+1}=\ln^*N/\Delta_t$ future value of the learning parameter.
\\
\noindent ENDFOR

\medskip\hrule\hrule\medskip
}
\smallskip

Let $A_T=\sum_{t=1}^T a_t$ be the loss of ConfHedge-2.
We keep the notation $H_T=\sum_{t=1}^T h_t$ and
$L^{(\q)}_T=\sum_{t=1}^T (\q_t\cdot\hat \l_t)$.
Theorem~\ref{main-result} also holds for these quantities.
Hence, using the same notation as in the Section~\ref{general-loss-1},
we obtain a bound (\ref{ineq-h-1}) and $H_T-L^{(\q)}_T\le\gamma_{k,T}\Delta_T$.

Since by convexity of the loss function
$a_t=\lambda(\omega_t,\gamma_t)=\lambda(\omega_t,\sum_{i=1}^N w_{i,t}\hat c_{i,t})\le
\sum_{i=1}^N w_{i,t}\hat l_{i,t}=h_t$ for all $t$, we have $A_T\le H_T$.

The confidence shifting regret of ConfHedge-2 is equal to
$$
R^{(\q)}_T=A_T-L^{(\q)}_T=\sum\limits_{t=1}^T\sum\limits_{i=1}^N q_{i,t}p_{i,t}(a_t-l_{i,t}).
$$
The upper bounds (\ref{main-result-1a}) of this regret are given by
Theorem~\ref{main-result} and by Corollary~\ref{main-result-1aw}.

\subsection{The electrical loads forecasting}\label{subsec-2}

The second group of numerical experiments were performed with the contest data of
the GefCom2012 competition conducted on the Kaggle platform (\citealt{GefCom2012}).
The main objective of this competition was to predict the daily course of hourly
electrical loads (demand values for electricity) in 20 regions according to
temperature records at 11 meteorological stations. Databases are available
at \url{http://www.kaggle.com/datasets}. The basic data were provided in the form of
the table ``temperature-history`` with archive records of temperature monitoring
at 11 meteorological stations and the table ``load-history`` with hourly
electrical load data recorded at 20 power distribution stations of the
region for the period from 01.01.2004 to 30.06.2008. The additional calendar
information (seasons, days of the week, and working days vs. holidays) could be also used.

As an illustration of how the proposed expert aggregation methods perform,
a simplified particular task was designed, namely, electrical load forecasting
in one of the power distribution networks (Zone5) one hour ahead based on
historical data and current calendar parameters. To account for temperature changes,
the temperature measurements of only one meteorological station (Station9) were
used; these data provided the best electrical load forecasts
in the selected network on the training part of the sample.

\begin{figure}[h!]
\centering\includegraphics[scale=0.50]
        %
	{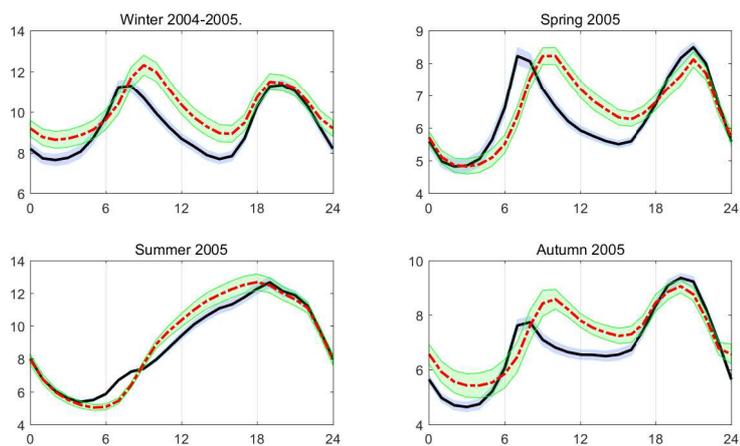}
          \caption{{\small Averaged curves of the daily electrical
loads for each of the four seasons of 2004-–2005. The color band around each curve
represents standard error of the mean.
The solid lines show the average level of electricity usage for working days,
and the dashed lines show the same estimates for the weekend days of
the same season}}\label{fig-2}
\end{figure}

Figure~\ref{fig-2} shows the averaged curves of the daily electrical
loads for each of the four seasons of 2004--2005 in the selected network.
We see that the course of the averaged curves clearly depends on the time of day,
and also varies from season to season. In addition, the working day and weekend
day patterns demonstrate distinct differences in the level of electricity usage.
Based on this figure, a simple scheme of forming an ensemble of experts, i.e.,
specialized algorithms that can only process strictly defined data, was chosen;
the scheme includes the following categories: four times of day
(night, morning, day, evening); working days and weekend days (two categories);
four seasons (winter, spring, summer, fall), all this giving $4\times 2\times 4=32$
specialized experts (Stepwise Linear Regression). We also use extra four experts,
each of which is focused on one of the seasons of the year, and one nonsleeping
expert (Random Forest algorithm). Thus, we used a total of 37 experts.

\begin{figure}[h!]
\centering\includegraphics[scale=0.55]
        {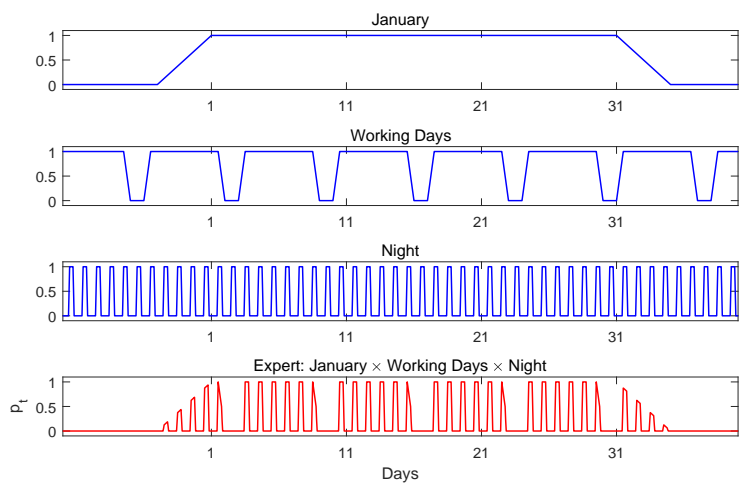}
         \caption{{\small Confidence function construction for the
expert January$\times$Working days$\times$Night}}\label{fig-3}
\end{figure}

At each moment of time, the confidence function of a given expert is calculated
as a product of the confidence functions for each of its specializations.
For example, Figure~\ref{fig-3} shows the stages of constructing the confidence
function for the expert focused on night forecasting (0--6 a.m.)
on the working days of January. Thus synthesized confidence functions are
used to form individual training samples for each expert at the stage of
training and to aggregate expert forecasts at the stage of testing.

To ensure a more smooth switch between experts, the membership functions
$p_{i,t}$ were formed as trapezoids, where the function takes the value 1
on the plateau corresponding to the selected calendar interval, and
varies linearly from 1 to 0 on the slopes. The slope width depends on the user
defined parameter.

At the stage of training, the following steps are taken for each algorithm (expert):
(1) For all elements of the training sample, a confidence level is calculated,
assuming its values are close to 1, if the sample is to be considered by the expert,
or close to 0, if the object is beyond its specialization. Based on
the confidence level, an individual training subsample is formed
for each algorithm from the full training sample.
(2) Based on this individual training subsample, a forecasting model is constructed.

\begin{figure}[h!]
\centering\includegraphics[scale=0.55]
          {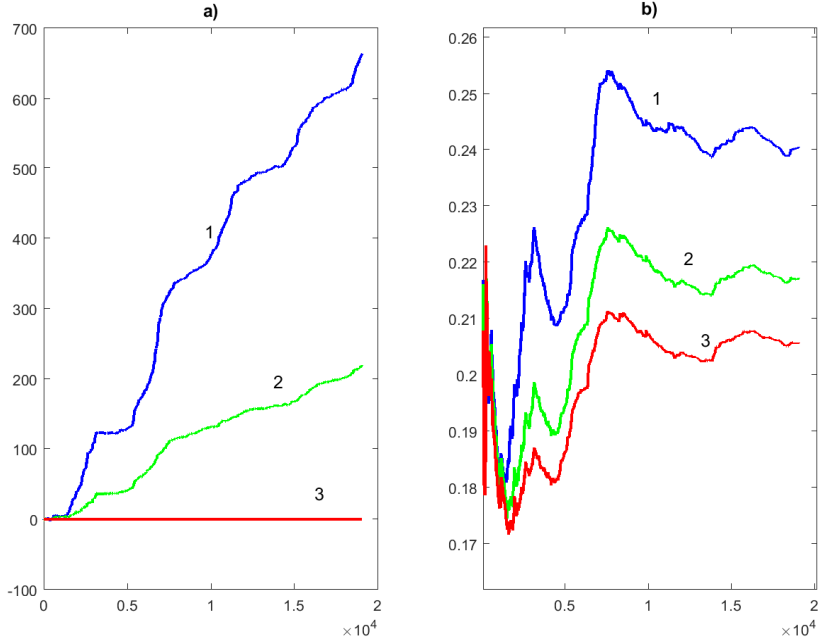}
          \caption{{\small
(a) The evolution of differences of cumulative losses
$L^1_T-L^3_T$ and $L^2_T-L^3_T$:
1 -- anytime nonsleeping expert (Random Forest algorithm),
2 -- ConfHedge-2 using ``sleeping experts'' model,
3 -- ConfHedge-2 using smooth confidence levels.
(b) The mean cumulative losses (MAE) of Random Forest (1) and
of two schemes of expert mixing (2 and 3)}}\label{fig-4}
\end{figure}

To compare the scheme of ``smooth mixing`` with the scheme of ``sleeping experts``,
the experiments on expert decision aggregation were performed in two stages.
First, only the scheme of mixing the sleeping and awake experts was used, i.e.,
the confidence level took only two values (0 or 1), and then the mixing algorithm
from Section 2 of this work was used.

The evolution of differences of cumulative losses $L^1_T-L^3_T$ and $L^2_T-L^3_T$,
where $L^1_T$ is the cumulative loss of anytime nonsleeping Random Forest algorithm and
$L^2_T$, $L^3_T$ are cumulative losses of two schemes of mixing
(``sleeping experts` and ``smooth mixing``), are shown in Figure~\ref{fig-4}a.

The mean cumulative losses (Mean Absolute Error -- MAE) $\frac{1}{T}L^1_T$ of Random
Forest algorithm and
of two schemes of expert mixing: $\frac{1}{T}L^2_T$ and $\frac{1}{T}L^3_T$,
are shown in Figure~\ref{fig-4}b.\footnote{The absolute loss function was used
in these experiments.}


In this experiment, the ``smooth mixing`` algorithm
outperforms the aggregating algorithm using ``sleeping experts`` and both these
algorithms outperform the anytime Random Forest forecasting algorithm.

\section{Conclusion}

In this paper we extend the AdaHedge algorithm by~\citet{follow} for a case
of shifting experts and for a smooth version of the method of specialized experts,
where at any time moment each expert's forecast is provided with a confidence level
which is a number between 0 and 1.

To aggregate experts predictions, we use methods of shifting experts and
the algorithm AdaHedge with an adaptive learning rate.
Due to this, we combine the advantages of both algorithms. We use the shifting regret
which is a more optimal characteristic of the algorithm, and we do not impose
restrictions on the expert losses. Also, we incorporate in this scheme
a smooth version of the method of
specialized experts by~\citet{BlM2007}, which allows us to make more flexible and
accurate predictions.

We obtained the new upper bounds for the regret of our algorithms, which generalize
similar upper bounds for the case of specialized experts.

A disadvantage of Theorem~\ref{main-result} and of Corollary~\ref{main-result-1aw}
is in asymmetry of the bounds (\ref{main-result-1aa-1}) and (\ref{main-result-1aa-3}) --
first of them has a term that depends quadratically on the
number $k$ of switches. Whether such a dependence is necessary is an open question.

All results are obtained in the adversarial setting, no assumptions are made
about the nature of data source.

We present the results of numerical experiments on short-term forecasting of electricity
consumption based on a real data. In these experiments, the ``smooth mixing`` algorithm
outperforms the aggregating algorithm with ``sleeping experts`` and both these
algorithms outperform the anytime Random Forest forecasting algorithm.

\appendix

\section{Main lemma}\label{tech-det-1}


For analysis of the mixing schemes, following~\citet{BoW2002},
we use the notion of relative entropy
$$
D(\p\|\q)=\sum\limits_{i=1}^n p_i\ln\frac{p_i}{q_i},
$$
where $n$ is an arbitrary positive integer number,
$\p=(p_1,\dots ,p_n)$, $\q=(q_1,\dots ,q_n)$ are elements of the $n$-dimensional simplex
of all probability distributions on a set of cardinality $n$. Put $0\ln 0=0$.

Consider some properties of the relative entropy.
The inequalities $\p>\q$, $\p\ge \q$, $\p\ge \0$ for vectors will be understood componentwise;
here $\0$ is the vector with zeros components.

\begin{lemma}\citealt{BoW2002}\label{relat-1}
For each $\p,\q,\w$ such that $\q,\w>\0$,
\begin{itemize}
\item
$
D(\p\|\q)\le D(\p\|\w)+\ln\left(\sum\limits_{i=1}^n p_i\frac{w_i}{q_i}\right).
$
\item
If $\q\ge r\w$ for some real number $r>0$ then
$
D(\p\|\q)\le D(p\|\w)+\ln\frac{1}{r}.
$
In particular, for $\p=\w$ we have  $D(\w\|\q)\le\ln\frac{1}{r}$ for each $\q\ge r\w$.
\item
Let $\p$ be a probability vector, $\q=\sum\limits_{i=0}^t\beta_i \w_i$,  where $\w_i>0$
for $0\le i\le t$, $\beta=(\beta_0,\dots ,\beta_t)$, $\sum_{i=0}^t\beta_i=1$,
and $\beta>\0$. Then
$
D(\p\|\q)\le D(\p\|\w_i)+\ln\frac{1}{\beta_i}
$
for each $i$.  In particular, if $\p=\q$ then
$
D\left(\w_i\|\sum\limits_{i=0}^n\beta_i \w_i\right)\le\ln\frac{1}{\beta_i}
$
for all $i$.
\end{itemize}
\end{lemma}
{\it Proof}. From concavity of the logarithm, we have
\begin{eqnarray}
D(\p\|\q)-D(\p\|\w)=\sum\limits_{i=1}^n p_i\ln\frac{w_i}{q_i}\le
\ln\left(\sum\limits_{i=1}^n p_i\frac{w_i}{q_i}\right).
\label{Kul-L-ineq-1}
\end{eqnarray}
If $\q\ge r\w$ then
$\sum\limits_{i=1}^n p_i\frac{w_i}{q_i}\le\sum\limits_{i=1}^n p_i\frac{w_i}{r w_i}=
\frac{1}{r}$.
$\qed$

The notion of mixloss
$
m_t=-\frac{1}{\eta_t}\sum\limits_{i=1}^N w_{i,t}e^{-\eta_t l_{i,t}}
$
and its cumulative variant $M_T=\sum\limits_{t=1}^T m_t$ are used in Hedge 
analisys.\footnote{
Mixloss is a very useful intermediate concept, cumulative variant of which
is less or equal to the cumulative loss of the best expert (up to a small term)
and, on the other hand, 
the cumulative mixloss is close to the cumulative loss of the aggregating algorithm.
For the logarithmic loss function, the mixloss coincides with the loss of the Vovk 
aggregating algorithm (see~\citealt{adams},~\citealt{cesa-bianchi},~\citealt{follow}).
}
By definition $m_t\le h_t$ for all $t$.

\begin{lemma}\label{Kul-L-ineq-3} (\citealt{BoW2002})
For any comparison vector $\q_t$,
\begin{eqnarray}
m_t-\sum_{i=1}^N q_{i,t}\hat l_{i,t}=\frac{1}{\eta_t}
\left(D(\q_t\|\w_t)-D(\q_t\|\w^\mu_t)\right).
\label{MPP-bound-1}
\end{eqnarray}
\end{lemma}
{\it Proof.} By (\ref{Kul-L-ineq-1}),
\begin{eqnarray}
m_t-\sum_{i=1}^N q_{i,t}\hat l_{i,t}=
\frac{1}{\eta_t}\sum\limits_{i=1}^N q_{i,t}\left(\ln e^{-\eta_t\hat l_{i,t}}-
\ln\sum\limits_{j=1}^N w_{j,t}e^{-\eta_t\hat l_{j,t}}\right)=
\nonumber
\\
\frac{1}{\eta_t}\sum\limits_{i=1}^N q_{i,t}\ln\frac{e^{-\eta_t\hat l_{i,t}}}
{\sum\limits_{j=1}^N w_{j,t}e^{-\eta_t\hat l_{j,t}}}=
\frac{1}{\eta_t}\sum\limits_{i=1}^N q_{i,t}\ln\frac{w^\mu_{i,t}}{w_{i,t}} =
\frac{1}{\eta_t}\left(D(\q_t\|\w_t)-D(\q_t\|\w^\mu_t)\right).~\qed
\nonumber
\end{eqnarray}

The following lemma presents a bound for the confidence regret in terms of cumulative
mixability gap.
\begin{lemma}\label{main-lemma-1}
Let $\alpha_t=\frac{1}{t}$ for all $t$ and the mixing scheme from Example 1 was used.
Then for any $T$, for any sequence of losses of the experts, and for any sequence of
comparison vectors $\q_t$ given on-line with no more than $k$ switches on the
time interval $1\le t\le T$,
\begin{eqnarray}
H_T-L^{(\q)}_T\le (k+2)(\ln T+1)\Delta_T.
\label{ineq-h-1}
\end{eqnarray}
\end{lemma}
{\it Proof}.
We apply Lemmas~\ref{relat-1} and~\ref{Kul-L-ineq-3}
for mixing schemes of Example 1 (Fixed Share).

Let a sequence $\l_t=(l_{1,t},\dots ,l_{N,t})$ of losses of the experts and a sequence
of comparison vectors $\q_t=(q_{1,t},\dots ,q_{N,t})$ be given on-line for $t=1,2,\dots$.
Assume that $T$ be an arbitrary and the comparison vector $\q_t$ changes
$k$ times for $1\le t\le T$.

We let $1<t_1<t_2<\dots t_k$ be the subsequence of indices in the sequence of
comparators $\q_1,\dots ,\q_T$, where shifting occurs:
$\q_{t_j}\not = \q_{t_j-1}$ and  $\q_t=\q_{t-1}$ for all other steps, where $t>1$.
Define also $t_0=1$ and $t_{k+1}=T+1$. We apply Lemma~\ref{Kul-L-ineq-3}
for the distribution $\beta^{t+1}$ from Example 1. Recall that
$w_{i,1}=w^\mu_{i,0}=\frac{1}{N}$ for $i=1,\dots , N$.

Summing (\ref{MPP-bound-1}) on time interval where $\q_t=\q_{t-1}$ for
$t_j+1\le t\le t_{j+1}-1$, we obtain
\begin{eqnarray}
\sum\limits_{t=t_j+1}^{t_{j+1}-1}\left(
m_t-\sum_{i=1}^N q_{i,t}\hat l_{i,t}
\right)
=\sum\limits_{t=t_j+1}^{t_{j+1}-1}
\frac{1}{\eta_t}\left(D(\q_t\|\w_t)-D(\q_t\|\w^\mu_t)\right)=
\nonumber
\\
\sum\limits_{t=t_j+1}^{t_{j+1}-1}
\left(\frac{1}{\eta_{t-1}}D(\q_t\|\w_t)-\frac{1}{\eta_t}D(\q_t\|\w^\mu_t)\right)+
\sum\limits_{t=t_j+1}^{t_{j+1}-1}
\left(\frac{1}{\eta_t}-\frac{1}{\eta_{t-1}}\right)D(\q_t\|\w_t)\le
\label{nom-1}
\\
\sum\limits_{t=t_j+1}^{t_{j+1}-1}
\left(\frac{1}{\eta_{t-1}}D(\q_t\|\w_t)-\frac{1}{\eta_t}D(\q_t\|\w^\mu_t)\right)+
\sum\limits_{t=t_j+1}^{t_{j+1}-1}\frac{1}{\ln^*N}\delta_{t-1}
\left(\ln N +\ln\frac{1}{\alpha_t}\right)\le
\label{nom-2}
\\
\sum\limits_{t=t_j+1}^{t_{j+1}-1}
\left(\frac{1}{\eta_{t-1}}D(\q_t\|\w^\mu_{t-1})-\frac{1}{\eta_t}D(\q_t\|\w^\mu_t)\right)+~~~~~~~~
\nonumber
\\
\sum\limits_{t=t_j+1}^{t_{j+1}-1}\frac{1}{\eta_{t-1}}\ln\frac{1}{1-\alpha_t}+
\sum\limits_{t=t_j+1}^{t_{j+1}-1}\left(\frac{1}{\ln^*N}\ln\frac{1}{\alpha_t}+1\right)
\delta_{t-1}\le~~~~~~~~
\label{nom-3}
\\
\frac{1}{\eta_{t_j}}D(\q_{t_j}\|\w^\mu_{t_j})-\frac{1}{\eta_{t_{j+1}-1}}
D(\q_{t_j}\|\w^\mu_{t_{j+1}-1})+~~~~~~~~
\nonumber
\\
\sum\limits_{t=t_j+1}^{t_{j+1}-1}\frac{1}{\eta_t}\ln\frac{1}{1-\alpha_t}+
\sum\limits_{t=t_j+1}^{t_{j+1}-1}\delta_{t-1}\ln\frac{1}{\alpha_t}+
\sum\limits_{t=t_j+1}^{t_{j+1}-1}\delta_{t-1}.~~~~~~~~
\label{nom-4}
\end{eqnarray}
In transition from (\ref{nom-1}) to (\ref{nom-2}), the inequality
$w_{i,t}\ge\frac{\alpha_t}{N}$ was used, then
\begin{eqnarray}
D(\q_t\|\w_t)=\sum\limits_{i=1}^N q_{i,t}\ln\frac{q_{i,t}}{w_{i,t}}\le
\sum\limits_{i=1}^N q_{i,t}\ln q_{i,t}-\sum\limits_{i=1}^N q_{i,t}\ln\frac{\alpha_t}{N}\le
\ln N+\ln\frac{1}{\alpha_t}.
\label{ex-3a}
\end{eqnarray}
In transition from (\ref{nom-2}) to (\ref{nom-3}), we use the inequality (\ref{Kul-L-ineq-1}),
where $s=t-1$,
$$
D(\q_t\|\w_t)\le D(\q_t\|\w^\mu_{t-1})+\ln\frac{1}{1-\alpha_t}.
$$
In transition from (\ref{nom-3}) to (\ref{nom-4}) the entropy terms within the sections
telescope and only for the beginning and the end of each section a positive and
a negative entropy term remains, respectively.
We also roughen the inequality (\ref{nom-3}) after division by $\ln^*N$.

For the beginnings of the $k$ sections $t=t_1,\dots ,t_k$ define
$s=0$, $\beta^{t_j}_0=\alpha_{t_j}$ in the inequality (\ref{Kul-L-ineq-1}), then
\begin{eqnarray}
m_{t_j}-\sum_{i=1}^N q_{i,{t_j}}\hat l_{i,{t_j}}\le
\frac{1}{\eta_{t_j}} D(\q_{t_j}\|\w^\mu_0)-
\frac{1}{\eta_{t_j}}D(\q_{t_j}\|\w^\mu_{t_j})+\frac{1}{\eta_{t_j}}\ln\frac{1}{\alpha_{t_j}}.
\label{ex-3b}
\end{eqnarray}
Summing all these inequalities and canceling out the corresponding terms, we obtain
\begin{eqnarray}
\sum_{t=1}^T
m_t-\sum_{i=1}^N q_{i,t}\hat l_{i,t}
\le
\sum\limits_{j=1}^k\left(\frac{1}{\eta_{t_j}}D(\q_{t_j}\|\w^\mu_0)-\frac{1}{\eta_{t_{j+1}-1}}
D(\q_{t_j}\|\w^\mu_{t_{j+1}-1})\right)+
\label{n0-4}
\\
\sum\limits_{t=2}^T\frac{1}{\eta_t}\ln\frac{1}{1-\alpha_t}+
\sum\limits_{t=2}^T\delta_{t-1}\ln\frac{1}{\alpha_t}+
\sum\limits_{t=2}^T\delta_{t-1}+
\sum\limits_{j=1}^k\Delta_{t_j-1}\ln\frac{1}{\alpha_{t_j}}\le
\label{n0-5}
\\
((k+2)\ln T+k+1)\Delta_T.
\label{n0-6}
\end{eqnarray}
In transition from (\ref{n0-4}) to (\ref{n0-5}) we use the inequality $D(\q\|\w^\mu_T)\ge 0$
for all $\q$ and equality $D(\q\|\w^\mu_0)=\ln N$. Then
$
\sum\limits_{j=1}^k\frac{1}{\eta_{t_j}}D(\q_{t_j}\|\w^\mu_0)\le k\Delta_T.
$
For $\alpha_t=\frac{1}{t}$ we use the inequality
\begin{eqnarray*}
\sum\limits_{t=2}^T\frac{1}{\eta_t}\ln\frac{1}{1-\alpha_t}\le\frac{1}{\eta_T}
\ln T\le\Delta_{T-1}\ln T\le\Delta_T\ln T.
\end{eqnarray*}

Since $H_T=M_T+\Delta_T$, the bound (\ref{n0-6}) implies (\ref{ineq-h-1}).
$\qed$

We will finish the proof of Theorem~\ref{main-result} at the end of Section~\ref{sec-3}.

The corresponding bounds for mixing scheme of Example 2 can be obtained in a similar way.
Since by definition $w_{i,t}\ge\frac{\alpha_t}{Nt}$ for each $t$,
the inequality (\ref{ex-3a}) is changed to
$D(\q_t\|\w_t)\le\ln N+\ln T+\ln\frac{1}{\alpha_t}$.
Also, the last term of the inequality (\ref{ex-3b}) is replaced by
$\frac{1}{\eta_{t_j}}\ln\frac{1}{t\alpha_{t_j}}$.
As a result, we obtain $\gamma_{k,T}=(2k+3)\ln T+(k+2)$.

In the case of bounded losses: $l_{i,t}\in [0,1]$, set in (\ref{nom-1}) -- (\ref{n0-6})
$\alpha_t=\frac{1}{t}$ and obtain
$M_T-L^{(\q)}_T\le\frac{1}{\eta_T}((k+2)\ln T+(k+1)\ln N)$.
Using the Hoeffding inequality $h_t\le m_t+\frac{\eta_t}{8}$,
where $\eta_t\sim\sqrt{\frac{\ln^*N}{t}}$, we obtain (\ref{bound-loss-1}).

\section{Technical bounds}\label{sec-3}

The derivation of the upper bound for $\Delta_T$ is similar to that given
in~\citet{follow}, except that the losses of experts are replaced
by $\hat l_{i,t}=E_{\p_{i,t}}[\tilde l_{i,t}]$.

Let $v_t=E_{j \sim \w_t} [(\hat l_{j,t}-E_{j\sim \w_t} [\hat l_{j,t}])^2]=
\sum\limits_{j=1}^N w_{j,t} (\hat l_{j,t}-h_t)^2$ and $V_T=\sum\limits_{t=1}^T v_t$.
\begin{lemma}\label{lemmm-1}
The quantity $\delta_t$ satisfies the inequality
\begin{eqnarray}\label{mix-ineq-21}
\delta_t \le \frac {e^{s_t \eta_t} - 1 - s_t \eta_t}{\eta_t s_t^2} v_t.
\end{eqnarray}
\end{lemma}
{\it Proof.}
The inequality (\ref{mix-ineq-21}) will be proved using Bernstein inequality
(see Lemmas 3-5 from~\citealt{cesa-bianchi}).
Let $X\in [0,1]$ be a random variable and $Var [X]$ be its variance.
Then for any $\eta>0$, we have $\ln E[e^{-\eta(X-E[X])}]\le Var [X](e^\eta-\eta-1)$.

Consider a random variable which takes the values $\hat l_{j,t}$ with probabilities
$w_{j,t}$, where $j = 1,\dots, N$.
Let us transform it so that its values belong to the segment $[0,1]$:
$X_t^j = \frac{\hat l_{j,t} - \hat l_t^-}{s_t}$.
Then the Bernstein inequality can be written as follows:
\begin{eqnarray}
\ln E_{j \sim \w_t}\left[ e^{-\eta \left( X_t^j - EX_t^j \right)} \right]
\le Var_{j \sim \w_t} [X_t^j]\left( e^{\eta} - 1 - \eta \right)
\label{ber-1}
\end{eqnarray}
for each $\eta>0$.
We rewrite this inequality in more detail for $\eta=s_t\eta_t$.

First, we simplify the left-hand side of the inequality (\ref{ber-1})
\begin{eqnarray*}
\ln E_{j\sim w_t}\left[ e^{-\eta \left( X_t^j - EX_t\right)} \right]
=\ln \left(\sum\limits_{j=1}^N w_{j,t}
e^{-s_t \eta_t \left(\frac{\hat l_{j,t} - l^-_t}{s_t} -
\sum\limits_{i=1}^N w_{i,t}\frac{\hat l_{i,t} - l^-_t}{s_t} \right)}\right) =
\nonumber
\\
\ln\left(\frac{\sum\limits_{j=1}^N w_{j,t} e^{-\eta_t\hat l_{j,t}}}
{e^{-\eta_t\sum\limits_{j=1}^N w_{j,t}\hat l_{j,t}}} \right) =
\ln \sum\limits_{j=1}^N w_{j,t} e^{-\eta_t\hat l_{j,t}} +
\eta_t\sum\limits_{j=1}^N w_{j,t}\hat l_{j,t} =
\eta_t(h_t-m_t)=\eta_t \delta_t.
\end{eqnarray*}
Then the inequality (\ref{ber-1}) can be written in the form
$
\eta_t \delta_t\le\frac{1}{s_t^2} v_t\left( e^{s_t \eta_t}-1-s_t\eta_t\right),
$
from which we obtain the required inequality (\ref{mix-ineq-21}).
$\qed$

The inequality (\ref{mix-ineq-21}) can be presented in the form
\begin{eqnarray}\label{mix-ineq-2}
\delta_t \le \frac{g(s_t\eta_t)}{s_t} v_t,\mbox{ where } g(x)=\frac{e^x-x-1}{x}.
\end{eqnarray}

\begin{lemma}\label{lemmm-2}
$\left( \Delta_T\right) ^2 \le (\ln^*N) V_T + \left(\frac{2}{3}\ln^*N+1\right) S_T\Delta_T$.
\end{lemma}
{\it Proof.} We have
\begin{eqnarray}
\left( \Delta_T \right) ^2 = \sum\limits_{t=1}^T \left((\Delta_t)^2 -
(\Delta_{t-1})^2 \right)=\sum\limits_{t=1}^T
\left( \left( \Delta_{t-1} + \delta_t \right)^2 - (\Delta_{t-1})^2 \right)\le
\nonumber
\\
\sum_{t=1}^T\left( 2\delta_t \Delta_{t-1} + \delta_t^2 \right) =
\sum\limits_{t=1}^T \left(\frac{2\delta_t}{\eta_t}\ln^*N +
\delta_t^2 \right) \le
\nonumber
\\
\sum\limits_{t=1}^T \left(\frac{2\delta_t}{\eta_t}\ln^*N + s_t
\delta_t \right) \le
2\ln^*N\sum\limits_{t=1}^T \frac{\delta_t}{\eta_t} + S_T \Delta_T.
\label{delta-1}
\end{eqnarray}
The bound for $\frac{\delta_t}{\eta_t}$ is obtained using (\ref{mix-ineq-2}):
$
\frac{1}{2}v_t\ge \frac{\delta_t s_t}{2 g \left( s_t \eta_t \right)}=
\frac{\delta_t}{\eta_t} -s_t\varphi(s_t\eta_t)\delta_t,
$
where $\varphi(x)=\frac{e^x - \frac 1 2 x^2 - x - 1}{xe^x - x^2 - x}$.
It is not difficult to prove that $\varphi(x)\le 1/3$.

Then, summing the inequality
$\frac{\delta_t}{\eta_t} \le \frac 1 3 s_t \delta_t + \frac 1 2 v_t$, and combining it
with the inequality (\ref{delta-1}) we obtain the needed inequality.
$\qed$

Now, we obtain the bounds for $V_T$. From definition
$v_t\le (l_t^+-h_t)(h_t-l_t^-)\le\frac{s^2_t}{4}$.

To obtain (\ref{main-result-1a}), we will use the following lemma.
\begin{lemma}\label{lll-1}
If $L^{(\q)}_T\le H_T$ then
$V_T\le S_T\frac{(L_T^+-L_T^{(\q)})(L_T^{(\q)}-L_T^-)}{L_T^+-L_T^-}+
\gamma_{k,T}S_T\Delta_T$.
\end{lemma}
{\it Proof.} The following inequality holds true
\begin{eqnarray}
V_T=\sum\limits_{t=1}^T v_t\le \sum\limits_{t=1}^T (l^+_t-h_t)(h_t-l^-_t)\le
S_T\sum\limits_{t=1}^T \frac{(l^+_t-h_t)(h_t-l^-_t)}{s_t}=
\nonumber
\\
S_T T \sum\limits_{t=1}^T \frac{1}{T}\frac{(l^+_t-h_t)(h_t-l^-_t)}{s_t} \le
S_T\frac{(L^+_T-H_T)(H_T-L^-_T)}{L^+_T-L^-_T}.
\label{vt-1a}
\end{eqnarray}
The inequality (\ref{vt-1a}) is obtained by applying the Jensen inequality
to a concave function
$B(x,y,z)=(z-y)(y-x)/(z-x)$ on the set $x\le y\le z$
(see for detail~\citealt{follow}).
$\qed$

Recall that $H_T\le L_T^{(\q)}+\gamma_{k,T}\Delta_T$.
Then, assuming $L_T^{(\q)}\le H_T$, we have
\begin{eqnarray}
V_T\le S_T\frac{(L^+_T-L^{(\q)}_T)(L^{(\q)}_T+\gamma_{k,T}\Delta_T-L^-_T)}{L^+_T-L^-_T}\le
\nonumber
\\
S_T\frac{(L^+_T-L^{(\q)}_T)(L^{(\q)}_T-L^-_T)}{L^+_T-L^-_T}+\gamma_{k,T}S_T\Delta_T.
\label{vt-1}
\end{eqnarray}
Denote
$$
Q_T=\frac{(L^+_T-L^{(\q)}_T)(L^{(\q)}_T-L^-_T)}{L^+_T-L^-_T}.
$$
By the inequality (\ref{vt-1}) and Lemma~\ref{lemmm-2}
\begin{eqnarray}
\Delta_T^2\le (S_T Q_T+\gamma_{k,T}S_T\Delta_T)\ln^*N
+\left(\frac{2}{3}\ln^*N+1\right)S_T\Delta_T=
\nonumber
\\
S_T Q_T\ln^*N+\left(\gamma_{k,T}\ln^*N+\frac{2}{3}\ln^*N+1\right)S_T\Delta_T.
\label{vvv-1}
\end{eqnarray}
We have the inequality $\Delta_T^2\le a+b\Delta_T$, where $a=S_T Q_T\ln^*N$,
$b=(\gamma_{k,T}\ln^*N+\frac{2}{3}\ln^*N+1)S_T$.

Solving this inequality with respect to $\Delta_T$, we obtain:
$
\Delta_T\le\frac{1}{2}b + \frac{1}{2}\sqrt{b^2 + 4a}\le
\sqrt{a} + b=\sqrt{S_T Q_T\ln^*N}+\left(\left(\gamma_{k,T}+\frac{2}{3}\right)\ln^*N+1\right)S_T.
$

If $H_T\le L^{(\q)}_T$ then $R^{(\q)}_T\le 0$ and the inequality (\ref{main-result-1a})
is automatically executed. Otherwise, by Lemma~\ref{main-lemma-1},
\begin{eqnarray*}
R^{(\q)}_T\le\gamma_{k,T}\Delta_T\le\gamma_{k,T}\sqrt{S_T Q_T\ln^*N}+
\gamma_{k,T}\left(\left(\gamma_{k,T}+\frac{2}{3}\right)\ln^*N+1\right)S_T.
\end{eqnarray*}
We obtain the inequality (\ref{main-result-1a}) using the lower and the upper
bounds (\ref{l-q-bounds-1}) for $L^{(\q)}_T$.

To obtain (\ref{main-result-1aa-4}), it is sufficient to use
the inequality $V_T\le\frac{1}{4}\sum\limits_{t=1}^T s^2_t$ and
a derivation similar to (\ref{vvv-1}).
This completes the proof of Theorem~\ref{main-result}.

To prove the inequality (\ref{main-result-1aa-3}) of Corollary~\ref{main-result-1aw}
we simplify the inequality (\ref{vt-1a}) as
$V_T\le S_T(L^+_T-H_T)\le S_T(L^+_T-L^{(q)}_T)$ if $L^{(q)}_T\le H_T$. Finally,
$
R^{(\q)}_T\le\gamma_{k,T}\Delta_T\le
\gamma_{k,T}\sqrt{S_T(L^+_T-L^{(q-)}_T)\ln^*N}+
\gamma_{k,T}\left(\frac{2}{3}\ln^*N+1\right)S_T
$
for all $T$.

\end{document}